\documentclass{article}

\usepackage{microtype}
\usepackage{graphicx}
\usepackage{booktabs}
\usepackage{hyperref}
\usepackage[preprint]{icml2026}
\usepackage{amsmath}
\usepackage{amssymb}
\usepackage{mathtools}
\usepackage{xcolor}
\usepackage{listings}
\lstdefinestyle{pseudocode}{
  basicstyle=\small\fontfamily{cmtt}\selectfont,
  keywordstyle=\bfseries,
  commentstyle=\itshape\color[rgb]{0.4,0.4,0.4},
  keywords={def,for,in,if,return},
  morecomment=[l]{\#},
  mathescape=true,
  frame=single,
  framesep=6pt,
  xleftmargin=4pt,
  xrightmargin=4pt,
  numbers=left,
  numberstyle=\tiny\color{gray},
  numbersep=6pt,
  aboveskip=8pt,
  belowskip=4pt,
  columns=fullflexible,
}
\usepackage[capitalize,noabbrev]{cleveref}
\graphicspath{{figures/}}

\icmltitlerunning{Conformal Risk Control for Safety-Critical Wildfire Prediction}

\begin{document}

\twocolumn[
    \icmltitle{Conformal Risk Control for Safety-Critical Wildfire Evacuation Mapping:\\
        A Comparative Study of Tabular, Spatial, and Graph-Based Models}

    \begin{icmlauthorlist}
        \icmlauthor{Baljinnyam Dayan}{imperial}
    \end{icmlauthorlist}
    \icmlaffiliation{imperial}{Imperial College London, London, United Kingdom}
    \icmlcorrespondingauthor{Baljinnyam Dayan}{baljinnyam.dayan25@imperial.ac.uk}
    % CID: 06059151
    \icmlkeywords{wildfire prediction, conformal risk control, safety-critical machine learning, uncertainty quantification}
    \vskip 0.3in
]
\printAffiliationsAndNotice{}

%% ============================================================
%% ABSTRACT
%% ============================================================

\begin{abstract}
    Every wildfire prediction model deployed today shares a dangerous property: none of these methods provides formal guarantees on how much fire spread is missed.
    Despite extensive work on wildfire spread prediction using deep learning, no prior study has applied distribution-free safety guarantees to this domain, leaving evacuation planners reliant on probability thresholds with no formal assurance.
    We address this gap by presenting, to our knowledge, the first application of conformal risk control (CRC) \citep{angelopoulos2022conformal} to wildfire spread prediction, providing finite-sample guarantees on false negative rate ($\mathrm{FNR} \le 0.05$).
    We expose a stark failure: across three model families of increasing complexity (tabular: LightGBM, AUROC 0.854; convolutional: Tiny U-Net, AUROC 0.969; and graph-based: Hybrid ResGNN-UNet, AUROC 0.964), standard thresholds capture only 7--72\% of true fire spread.
    CRC eliminates this failure uniformly.
    Our central finding is that \emph{model architecture determines evacuation efficiency, while CRC determines safety}: both spatial models with CRC achieve ${\sim}95\%$ fire coverage while flagging only ${\sim}15\%$ of pixels of total map, making them 4.2$\times$ more efficient than LightGBM, while the graph model's additional complexity over a simple U-Net yields no meaningful efficiency gain.
    We propose a shift-aware three-way CRC framework that assigns SAFE/MONITOR/EVACUATE zones for operational triage, and characterize a fundamental limitation of prevalence-weighted bounds under extreme class imbalance (${\sim}5\%$ fire prevalence).
    All models, calibration code, and evaluation pipelines are released for reproducibility.
\end{abstract}

%% ============================================================
%% 1. INTRODUCTION
%% ============================================================

\section{Introduction}

Wildfire evacuation planning requires prediction systems that prioritize safety over accuracy.
In this domain, a false negative (a fire pixel classified as safe) poses direct risk to human life, while false positives carry their own operational cost: overly broad evacuation perimeters trigger \emph{shadow evacuations}, where residents outside the official zone self-evacuate, congesting road networks and potentially trapping those in genuine danger \citep{zhao2022evacuation}.
This dual asymmetry renders standard accuracy-based evaluation inadequate: a system reporting 90\% pixel-level accuracy may simultaneously miss the majority of fire pixels due to extreme class imbalance, as we demonstrate empirically in this work.
Controlling pixel-level FNR is a necessary condition for safe evacuation mapping: a system that misses individual fire pixels cannot produce reliable fire perimeters, even if downstream spatial post-processing is applied.

Despite decades of wildfire prediction research \citep{jain2020review} and the recent availability of distribution-free safety tools such as \emph{conformal risk control} (CRC) \citep{angelopoulos2022conformal}, no prior work has applied CRC, or any conformal method, to wildfire spread prediction.
Wildfire ML continues to evaluate models on accuracy-based metrics (F1, IoU, AUROC) that are blind to the safety-critical nature of the task, while the conformal prediction community has focused on medical imaging, autonomous driving, and natural language processing.
This leaves a clear gap: \textbf{the domain where false negatives are most dangerous has no formal safety guarantees.}

We bridge this gap.
CRC selects a decision threshold from held-out calibration data such that $\mathbb{E}[\mathrm{FNR}] \le \alpha$ on test data, regardless of the underlying model's quality or calibration \citep{angelopoulos2021gentle}.
A natural question then motivates our study: \emph{can a sufficiently powerful model eliminate the need for post-hoc safety calibration?}
To answer this, we train and evaluate three architectures of increasing complexity on the Next Day Wildfire Spread (NDWS) dataset \citep{huot2022next}:
\textbf{(i)}~a tabular LightGBM baseline \citep{ke2017lightgbm} treating each pixel independently,
\textbf{(ii)}~a Tiny U-Net \citep{ronneberger2015unet} capturing local spatial structure, and
\textbf{(iii)}~a Hybrid ResGNN-UNet combining convolutional encoders with graph attention layers \citep{veličković2018graph} to model fire propagation on pixel graphs.
All models are trained from scratch on identical data splits.

Our experiments yield a clear answer: \textbf{no model is safe without CRC}.
At their respective optimal thresholds, LightGBM misses 92.8\% of fires, U-Net misses 61.0\%, and the ResGNN-UNet misses 28.5\%.
With CRC, all three models achieve ${\ge}94\%$ coverage, but spatial models flag only ${\sim}15\%$ of pixels compared to LightGBM's 62.6\%, confirming that spatial inductive bias drives efficiency under the same safety guarantee.

\begin{figure*}[t]
    \centering
    \includegraphics[width=0.85\textwidth]{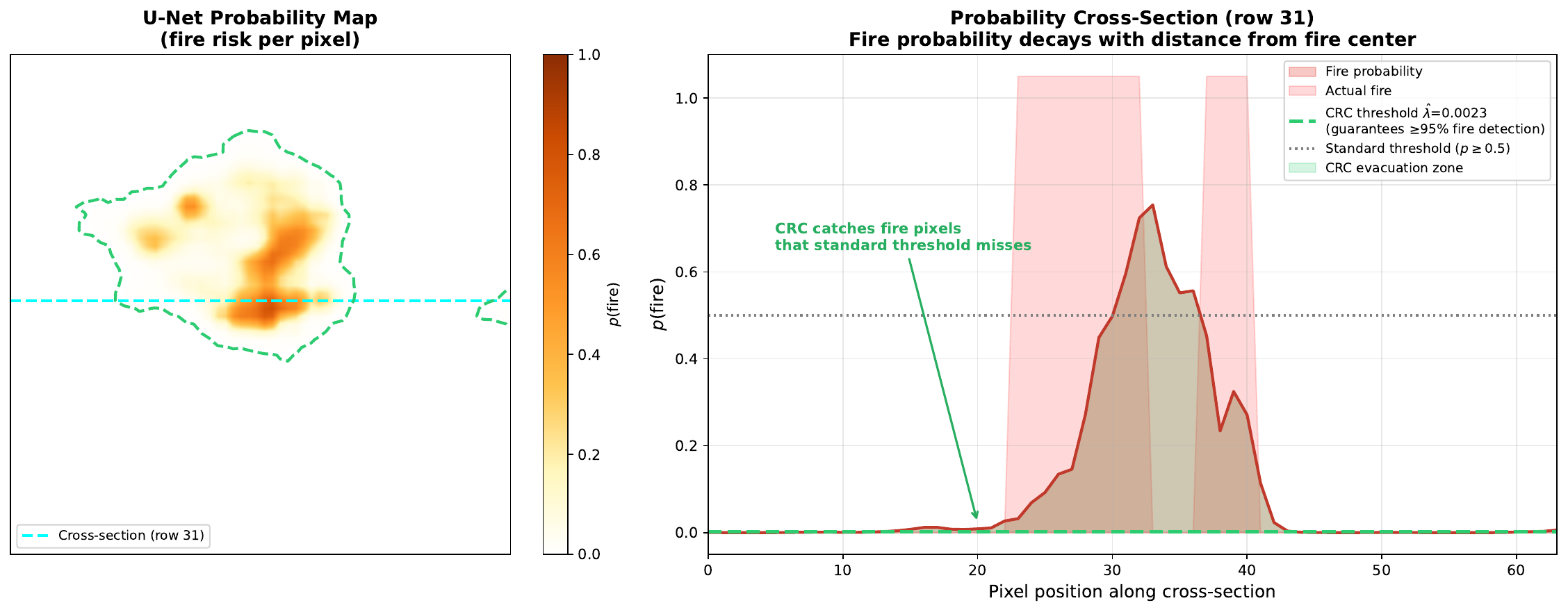}
    \caption{Why standard thresholds fail.  A cross-section through a U-Net fire prediction shows probability peaking at ${\sim}0.8$ and decaying smoothly.  The standard threshold ($\hat{p} \ge 0.5$, gray) misses the majority of the fire region; the CRC threshold ($\hat{\lambda} = 0.002$, green) captures it entirely.}
    \label{fig:cross-section}
\end{figure*}

\paragraph{Contributions.}
\begin{itemize}
    \item We demonstrate that standard probability thresholding is \emph{fundamentally inadequate} for wildfire evacuation across three model families, with fire coverage ranging from 7\% to 72\%.
    \item We show that CRC provides guaranteed ${\ge}95\%$ fire coverage regardless of model quality, and that architecture determines \emph{efficiency} (evacuation zone size), with spatial models achieving 4.2$\times$ tighter zones than tabular baselines under the same safety guarantee.
    \item We propose shift-aware three-way CRC with SAFE/MONITOR/EVACUATE zones and characterize its behavior under extreme class imbalance, identifying a fundamental limitation of prevalence-weighted bounds when $\pi_0 \gg \pi_1$.
    \item We provide a complete, reproducible pipeline with deterministic splits and open-source code for all experimental phases.\footnote{\url{https://github.com/baljinnyamday/wildfire-evacuation-crc}}
\end{itemize}

%% ============================================================
%% 2. RELATED WORK
%% ============================================================

\section{Related Work}

\paragraph{Machine learning for wildfire prediction.}
Machine learning has been widely applied to wildfire risk assessment and spread prediction \citep{jain2020review, radke2019firecast}.
The NDWS dataset \citep{huot2022next} established a standardized pixel-level benchmark, enabling systematic comparison across architectures including graph neural networks for spatial propagation \citep{kipf2017semi, veličković2018graph}.
However, \emph{every} existing work evaluates exclusively on accuracy-based metrics (F1, IoU, AUROC); none provides formal safety guarantees on the false negative rate.

\paragraph{Conformal prediction and distribution-free inference.}
Conformal prediction \citep{vovk2005algorithmic, shafer2008tutorial} provides distribution-free prediction sets with finite-sample coverage guarantees under exchangeability.
\citet{romano2020classification} extended these ideas to adaptive classification, and \citet{angelopoulos2022conformal} generalized them to \emph{conformal risk control} (CRC), which controls arbitrary monotone loss functions---including the FNR loss central to our work.
CRC has been applied in medical imaging and autonomous driving, yet geospatial hazard prediction remains unexplored.
To our knowledge, \textbf{this is the first application of CRC to wildfire spread prediction}, where the asymmetric cost structure---missed fires endanger lives; over-broad zones cause shadow evacuations \citep{zhao2022evacuation}---makes formal FNR guarantees uniquely valuable.

\paragraph{Uncertainty quantification and calibration.}
Modern neural networks are poorly calibrated \citep{guo2017calibration}, and standard recalibration methods such as Platt scaling \citep{platt1999probabilistic} and temperature scaling improve confidence estimates but provide no finite-sample coverage guarantees.
We show that even well-calibrated models remain unsafe under standard thresholding due to the extreme class imbalance in wildfire data (${\sim}5\%$ fire pixels), motivating distribution-free bounds rather than approximate confidence.

\paragraph{Cost-sensitive learning and spatial extremes.}
Focal loss \citep{lin2017focal} and cost-sensitive learning \citep{elkan2001foundations} improve empirical recall under imbalance but provide no finite-sample FNR bound.
Neyman--Pearson classification \citep{scott2005neyman} targets a type-II error constraint with only asymptotic guarantees.
Spatial extreme-value theory \citep{coles2001introduction, davison2012statistical} models tail behavior of spatial processes but requires parametric assumptions unsuited to pixel-level spread prediction.
CRC is complementary: distribution-free, finite-sample, and applicable post-hoc to any model---including those already trained with cost-sensitive losses.

\paragraph{Safety-critical machine learning.}
In safety-critical domains, the cost of false negatives far exceeds that of false positives, motivating formal guarantees over empirical performance alone \citep{amodei2016concrete}.
Wildfire evacuation is a stark example---yet it is precisely the domain where such guarantees have been absent.
Our work closes this gap, demonstrating a 23--87 percentage-point improvement in fire detection rate when replacing standard thresholds with CRC-calibrated ones.

%% ============================================================
%% 3. PRELIMINARIES
%% ============================================================

\section{Preliminaries}
\label{sec:prelim}

\subsection{Problem Formulation}

We consider pixel-level binary classification for wildfire spread prediction.
Given an input sample $x \in \mathbb{R}^{H \times W \times C}$ (in our setting, $H{=}W{=}64$, $C{=}12$) and binary fire mask $y \in \{0,1\}^{H \times W}$, a model $f_\theta$ outputs per-pixel probability estimates $\hat{p}_{ij} = f_\theta(x)_{ij} \in [0,1]$.
Binary predictions are obtained by thresholding:
\begin{equation}
    \hat{y}_{ij}(\lambda) = \mathbb{1}\{\hat{p}_{ij} \ge \lambda\}.
\end{equation}
Our primary safety objective is to control the false negative rate:
\begin{equation}
    \mathrm{FNR}(\lambda) = \frac{\sum_{ij} y_{ij}(1 - \hat{y}_{ij}(\lambda))}{\sum_{ij} y_{ij}} \le \alpha, \quad \alpha = 0.05,
    \label{eq:fnr}
\end{equation}
equivalently requiring empirical fire coverage $1 - \mathrm{FNR} \ge 95\%$.

\subsection{Conformal Risk Control}
\label{sec:crc-background}

CRC \citep{angelopoulos2022conformal} provides a principled method for selecting $\lambda$ with finite-sample guarantees.
Given calibration examples $\{(x_i, y_i)\}_{i=1}^{n}$ drawn exchangeably with the test data, let $m$ denote the number of positive (fire) calibration pixels and $p_{(1)} \le \cdots \le p_{(m)}$ their sorted predicted probabilities.  CRC selects:
\begin{equation}
    \hat{\lambda} = p_{(\lceil \alpha\,(m+1) \rceil)},
    \label{eq:crc}
\end{equation}
the $\alpha$-quantile of positive-pixel probabilities with finite-sample correction.
At threshold $\hat{\lambda}$, at most $\lceil \alpha(m{+}1) \rceil - 1$ positive pixels fall below it, controlling the empirical FNR $\le \alpha$.
The resulting guarantee is:
\begin{equation}
    \mathbb{E}[\mathrm{FNR}(\hat{\lambda})] \le \alpha,
    \label{eq:guarantee}
\end{equation}
which holds \emph{regardless} of the model's quality, calibration, or architecture, requiring only exchangeability of calibration and test data.

\paragraph{Exchangeability assumption.}
The guarantee in \Cref{eq:guarantee} requires that calibration and test samples are drawn exchangeably from the same distribution.
Our pixel-pooled formulation treats pixels across different images as exchangeable units; in practice, pixels within the same $64{\times}64$ patch exhibit spatial correlation, which is a standard simplification in conformal segmentation tasks.
We discuss the implications of this assumption, including potential violations under temporal and geographic shift, in \Cref{sec:limitations}.

\begin{figure}[t]
    \centering
    \includegraphics[width=\columnwidth]{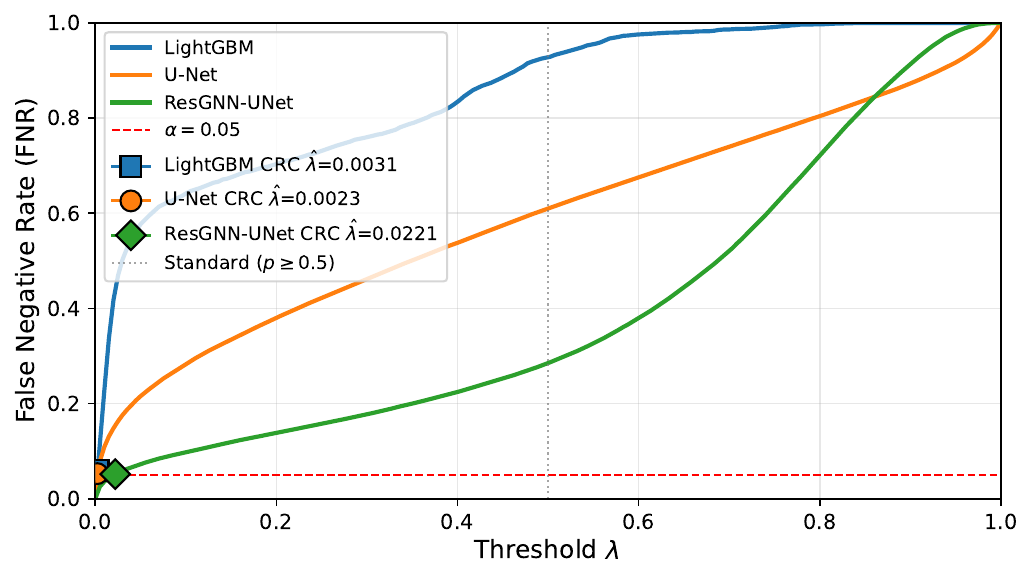}
    \caption{FNR as a function of decision threshold for all three models.  The red dashed line marks $\alpha = 0.05$.  Both spatial models maintain safe FNR over a much wider threshold range than LightGBM, explaining their 4.2$\times$ smaller evacuation zones under CRC.  The near-overlapping U-Net and ResGNN-UNet curves illustrate diminishing returns from architectural complexity.  Markers show each model's CRC $\hat{\lambda}$.}
    \label{fig:fnr-sweep}
\end{figure}

\paragraph{Safety--efficiency decomposition.}
A key insight motivating our study is that CRC \emph{decouples} safety from efficiency.
\Cref{eq:guarantee} holds for any model; what varies is the \emph{prediction set size}, the fraction of pixels flagged as fire.
A model with strong fire/no-fire separation allows CRC to set a higher $\hat{\lambda}$, flagging fewer pixels while maintaining the same safety guarantee.
A poorly separating model forces $\hat{\lambda}$ toward zero, expanding the evacuation zone.
Thus, the model determines efficiency; CRC determines safety.
\Cref{fig:fnr-sweep} illustrates this: both spatial models maintain safe FNR over a much wider threshold range than LightGBM, explaining the large difference in evacuation zone size when all three are wrapped in CRC.

%% ============================================================
%% 4. METHODS
%% ============================================================

\section{Methods}

We train three architectures of increasing complexity on the same dataset and splits, then apply CRC uniformly.

\subsection{Data}

We use the NDWS dataset \citep{huot2022next, ndwsdataset}: 18,545 samples of $64{\times}64$ patches with 12 input channels and a binary fire target (\Cref{tab:features}).
We enforce a deterministic split (seed 42): \textbf{train} 12,981 (70\%), \textbf{calibration} 2,781 (15\%), \textbf{test} 2,783 (15\%).
The calibration set is never used during model training.
Fire targets include no-data pixels ($-1$); we construct valid-pixel masks and apply masked loss and evaluation throughout.
Only ${\sim}5\%$ of valid pixels are labeled fire, creating extreme class imbalance with profound implications for threshold selection (\Cref{sec:results}).

\begin{table}[t]
    \caption{NDWS input features (12 channels per $64{\times}64$ pixel patch).  All features are from remote-sensing and reanalysis sources \citep{huot2022next}.}
    \label{tab:features}
    \centering
    \footnotesize
    \setlength{\tabcolsep}{3pt}
    \begin{tabular}{llll}
        \toprule
        Category     & Feature      & Variable             & Unit                      \\
        \midrule
        Topography   & elevation    & Terrain height       & m                         \\
        \midrule
        Weather      & th           & Wind direction       & $^\circ$                  \\
                     & vs           & Wind speed           & m/s                       \\
                     & tmmn         & Min temperature      & K                         \\
                     & tmmx         & Max temperature      & K                         \\
                     & sph          & Specific humidity    & kg/kg                     \\
                     & pr           & Precipitation        & mm                        \\
        \midrule
        Drought      & pdsi         & Palmer Drought Index & ---                       \\
        Fire weather & erc          & Energy Release Comp. & ---                       \\
        Vegetation   & NDVI         & Vegetation index     & $[-1,1]$                  \\
        Demographics & population   & Population density   & ppl/km\textsuperscript{2} \\
        Prior state  & PrevFireMask & Previous fire mask   & $\{0,1\}$                 \\
        \bottomrule
    \end{tabular}
\end{table}

\subsection{Model 1: Tabular Baseline (LightGBM)}

We flatten each $64{\times}64$ patch into pixel-level rows with 12 features and train an LGBMClassifier \citep{ke2017lightgbm} with binary cross-entropy objective.
To control memory, we subsample 5\% of training pixels.
This model treats each pixel independently and serves as a baseline to isolate the contribution of spatial inductive biases.

\subsection{Model 2: Tiny U-Net}

We implement a shallow U-Net \citep{ronneberger2015unet} with two encoder blocks ($12{\to}32{\to}64$ channels), a bottleneck ($64{\to}128$), and two decoder blocks with skip connections (${\sim}470$K parameters).
A final $1{\times}1$ convolution with sigmoid activation produces per-pixel probabilities.
Training uses AdamW (lr$=10^{-3}$, weight decay$=10^{-4}$) with cosine annealing over 50 epochs, masked BCE loss, and best-model checkpointing on validation loss.

\subsection{Model 3: Hybrid ResGNN-UNet}
\label{sec:resgnn}

To test whether more expressive architectures improve probability separation, we implement a Hybrid ResGNN-UNet (${\sim}229$K parameters):
\textbf{(i)}~a three-block CNN encoder ($12{\to}32{\to}64{\to}64$) downsampling to $8{\times}8$;
\textbf{(ii)}~a graph attention bottleneck with three GATConv layers \citep{veličković2018graph} (4 heads each) on an 8-connected pixel grid, with a residual connection;
\textbf{(iii)}~a U-Net decoder with three upsampling blocks, skip connections, and Dropout2d.
Training uses AdamW (lr$=3{\times}10^{-4}$, weight decay$=10^{-3}$) with BCEWithLogitsLoss (pos\_weight$=8.0$), cosine annealing over 25 epochs, gradient clipping, and early stopping (patience$=$7).

The GNN bottleneck treats the downsampled feature map as a graph where each pixel attends to its 8 spatial neighbors, designed to capture fire propagation patterns.

\subsection{CRC Calibration}
\label{sec:crc-method}

For each model, we compute per-pixel probabilities on the calibration set, then select $\hat{\lambda}$ via \Cref{eq:crc} with $\alpha = 0.05$.
This threshold is applied unchanged to the test set.

\subsection{Three-Way CRC Extension}
\label{sec:threeway}

We adapt shift-aware CRC to wildfire evacuation, routing each pixel to one of three actions:
\begin{equation}
    \begin{aligned}
        \text{SAFE} \;     & :\; \hat{p} < \lambda_{\min},                    \\
        \text{MONITOR} \;  & :\; \lambda_{\min} \le \hat{p} < \lambda_{\max}, \\
        \text{EVACUATE} \; & :\; \hat{p} \ge \lambda_{\max}.
    \end{aligned}
\end{equation}
The MONITOR zone routes uncertain pixels to human review rather than forcing a binary automated decision.
We use cost-weighted CRC with $c_{\mathrm{fn}}{=}5$, $c_{\mathrm{fp}}{=}1$ \citep{zhao2022evacuation} and cost-weighted risk level $\alpha_{\mathrm{cw}}{=}0.50$ (cost-weighting with ${\sim}95\%$ non-fire pixels shifts the risk scale relative to the binary FNR target).

To account for seasonal prevalence variation, we specify a shift interval $[\rho_{\mathrm{lo}}, \rho_{\mathrm{hi}}] = [0.9, 1.1]$ on the prevalence ratio $\rho = \pi_1^{\mathrm{dep}} / \pi_1^{\mathrm{cal}}$.
The adjusted calibration level is:
\begin{align}
    \epsilon_{\max}        & = B_{\mathrm{pw}} \cdot \max\!\bigl(\lVert \boldsymbol{\delta}_{\rho_{\mathrm{lo}}} \rVert_1,\, \lVert \boldsymbol{\delta}_{\rho_{\mathrm{hi}}} \rVert_1\bigr) + \tfrac{B_{\mathrm{pw}}}{N{+}1}, \label{eq:eps-max} \\
    \alpha_{\mathrm{safe}} & = \alpha_{\mathrm{cw}} - \epsilon_{\max}, \label{eq:alpha-safe}
\end{align}
where $B_{\mathrm{pw}} = \max(c_{\mathrm{fn}} \pi_1,\, c_{\mathrm{fp}} \pi_0)$ is the prevalence-weighted bound and $\boldsymbol{\delta}_\rho$ the importance weight mismatch at shift $\rho$.
A base threshold $\hat{\lambda}$ is found via cost-weighted CRC at level $\alpha_{\mathrm{safe}}$, then shifted proportionally to the endpoint weight mismatches to produce $\lambda_{\min}$ and $\lambda_{\max}$ (Algorithm~\ref{alg:threeway}).
On the decided set, $\mathbb{E}[\ell \mid \mathrm{decided}] \le \alpha_{\mathrm{cw}} / (1 - d)$, where $d$ is the MONITOR fraction.
If $\alpha_{\mathrm{safe}} \le 0$, the interval is too wide for safe calibration.

%% ============================================================
%% 5. EXPERIMENTS
%% ============================================================

\section{Experiments}
\label{sec:results}

\subsection{Setup}

We evaluate three model families under three decision frameworks:
\textbf{(i)}~standard thresholding at $\hat{p} \ge 0.5$,
\textbf{(ii)}~CRC with $\alpha = 0.05$, and
\textbf{(iii)}~three-way CRC with cost-weighted calibration.
All thresholds are computed on the held-out calibration set and applied unchanged to the test set.

\paragraph{Metrics.}
Our \emph{primary} metrics are:
\textbf{coverage} $(1{-}\mathrm{FNR})$, the fraction of true fire pixels correctly flagged (target: ${\ge}95\%$); and
\textbf{set size}, the fraction of all valid pixels flagged as fire (smaller is better, given safety is met).
\emph{Secondary} metrics include AUROC, precision, F1, IoU, AUPRC, and three-way zone statistics.

\subsection{Main Results}

\begin{table}[t]
    \caption{Test-set results across three models and three decision frameworks.  CRC targets $\mathrm{FNR} \le 0.05$.  Three-way CRC uses cost-weighted $\alpha_{\mathrm{cw}}{=}0.50$, $c_{\mathrm{fn}}{=}5$, $c_{\mathrm{fp}}{=}1$, $\rho \in [0.9, 1.1]$.  All models trained from scratch on identical 70/15/15 splits.  Values show $\pm$ 95\% bootstrap CI (10K image resamples).}
    \label{tab:main-results}
    \centering
    \footnotesize
    \setlength{\tabcolsep}{2.5pt}
    \resizebox{\columnwidth}{!}{%
        \begin{tabular}{lcccccr}
            \toprule
            Method              & Params        & $\hat{\lambda}$      & Cov.$\uparrow$               & FNR$\downarrow$              & Set$\downarrow$              & AUC$\uparrow$ \\
            \midrule
            LGBM ($p{\ge}.5$)   & ---           & .500                 & .072{\tiny$\pm$.02}          & .928{\tiny$\pm$.02}          & .001                         & .854          \\
            LGBM+CRC            & ---           & .003                 & .941{\tiny$\pm$.01}          & .059{\tiny$\pm$.01}          & .626{\tiny$\pm$.01}          & .854          \\
            LGBM+3way           & ---           & {\scriptsize[0,.03]} & 1.00                         & .000                         & .040                         & .854          \\
            \midrule
            UNet ($p{\ge}.5$)   & 470K          & .500                 & .390{\tiny$\pm$.02}          & .610{\tiny$\pm$.02}          & .007                         & .969          \\
            \textbf{UNet+CRC}   & \textbf{470K} & \textbf{.002}        & \textbf{.947}{\tiny$\pm$.01} & \textbf{.053}{\tiny$\pm$.01} & \textbf{.149}{\tiny$\pm$.01} & \textbf{.969} \\
            UNet+3way           & 470K          & {\scriptsize[0,.02]} & 1.00                         & .000                         & .051                         & .969          \\
            \midrule
            ResGNN ($p{\ge}.5$) & 229K          & .500                 & .715{\tiny$\pm$.02}          & .285{\tiny$\pm$.02}          & .031{\tiny$\pm$.00}          & .964          \\
            ResGNN+CRC          & 229K          & .022                 & .948{\tiny$\pm$.01}          & .052{\tiny$\pm$.01}          & .151{\tiny$\pm$.01}          & .964          \\
            ResGNN+3way         & 229K          & {\scriptsize[0,.02]} & 1.00                         & .000                         & .149                         & .964          \\
            \bottomrule
        \end{tabular}}
\end{table}

\Cref{tab:main-results} reveals four key findings.

\paragraph{Standard thresholds are unsafe.}
At $\hat{p} \ge 0.5$, LightGBM captures only 7.2\% of fire pixels, U-Net 39.0\%, and ResGNN-UNet 71.5\%.
Even the best-performing model at a standard threshold misses nearly a third of all fires.
The ${\sim}95\%$ non-fire class dominance concentrates probability mass near zero, making any conventional threshold unsafe (\Cref{fig:cross-section}).
The FNR sweep (\Cref{fig:fnr-sweep}) confirms this: safe FNR is achievable only at thresholds far below 0.5.

\begin{figure}[t]
    \centering
    \includegraphics[width=\columnwidth]{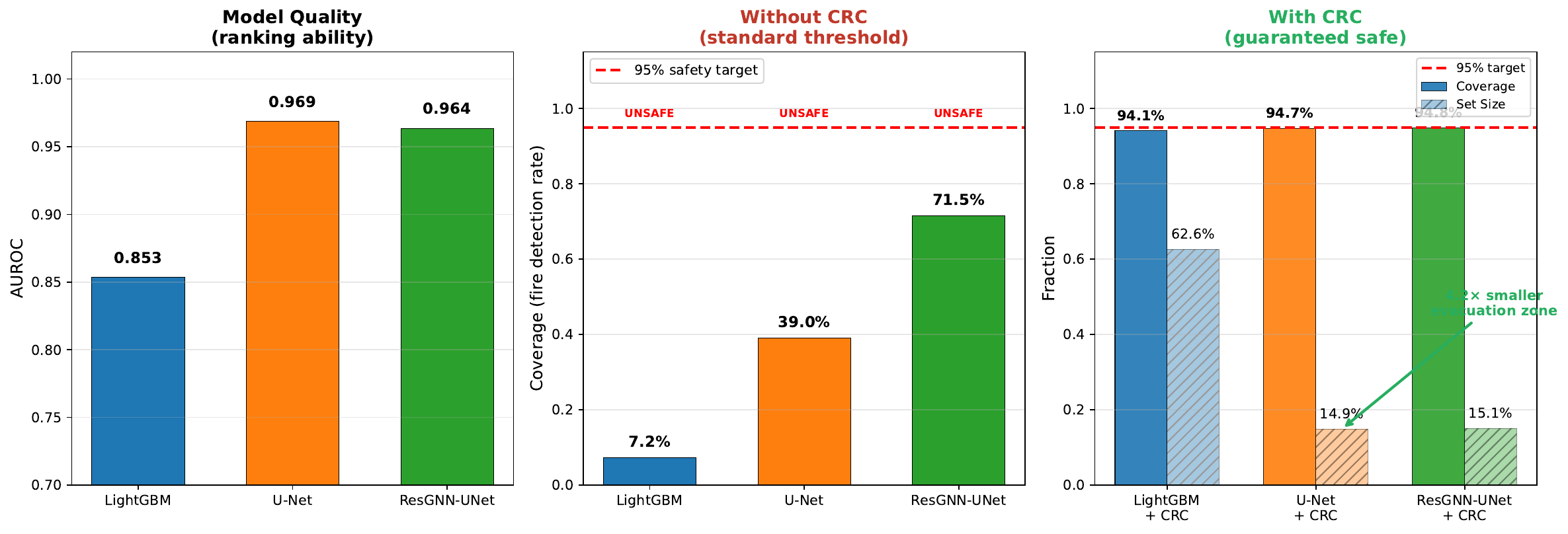}
    \caption{Three-model comparison.  \textbf{Left:} AUROC.  \textbf{Center:} without CRC, no model reaches 95\% coverage.  \textbf{Right:} with CRC, all three meet the safety target; both spatial models achieve ${\sim}4\times$ smaller evacuation zones than LightGBM.}
    \label{fig:three-model}
\end{figure}

\paragraph{CRC delivers safety; model quality delivers efficiency.}
CRC raises all three models to ${\ge}94\%$ coverage, meeting the $\mathrm{FNR} \le 0.05$ target.
However, the cost differs substantially (\Cref{fig:three-model}):
LightGBM must flag 62.6\% of all pixels, while both spatial models flag only ${\sim}15\%$, a \textbf{4.2$\times$ reduction} in evacuation zone size.
This gap reflects the AUROC difference between tabular and spatial models (0.854 vs.\ 0.964--0.969): better fire/non-fire separation allows CRC to set a higher $\hat{\lambda}$ while maintaining the same safety guarantee.
\Cref{fig:safety-efficiency} (Appendix) visualizes this safety--efficiency decomposition across all nine configurations.

\paragraph{Spatial inductive bias, not graph complexity, drives efficiency.}
The ResGNN-UNet (AUROC 0.964) nearly matches the Tiny U-Net (0.969), and both achieve near-identical CRC efficiency: 15.1\% vs.\ 14.9\% set size under the same safety guarantee.
On $64{\times}64$ patches, the U-Net's receptive field already covers the full spatial extent; the GNN bottleneck adds engineering complexity without meaningful improvement.
This reinforces our central claim: \emph{CRC determines safety, spatial inductive bias determines efficiency, and additional architectural complexity yields diminishing returns}.

\begin{figure*}[t]
    \centering
    \includegraphics[width=0.85\textwidth]{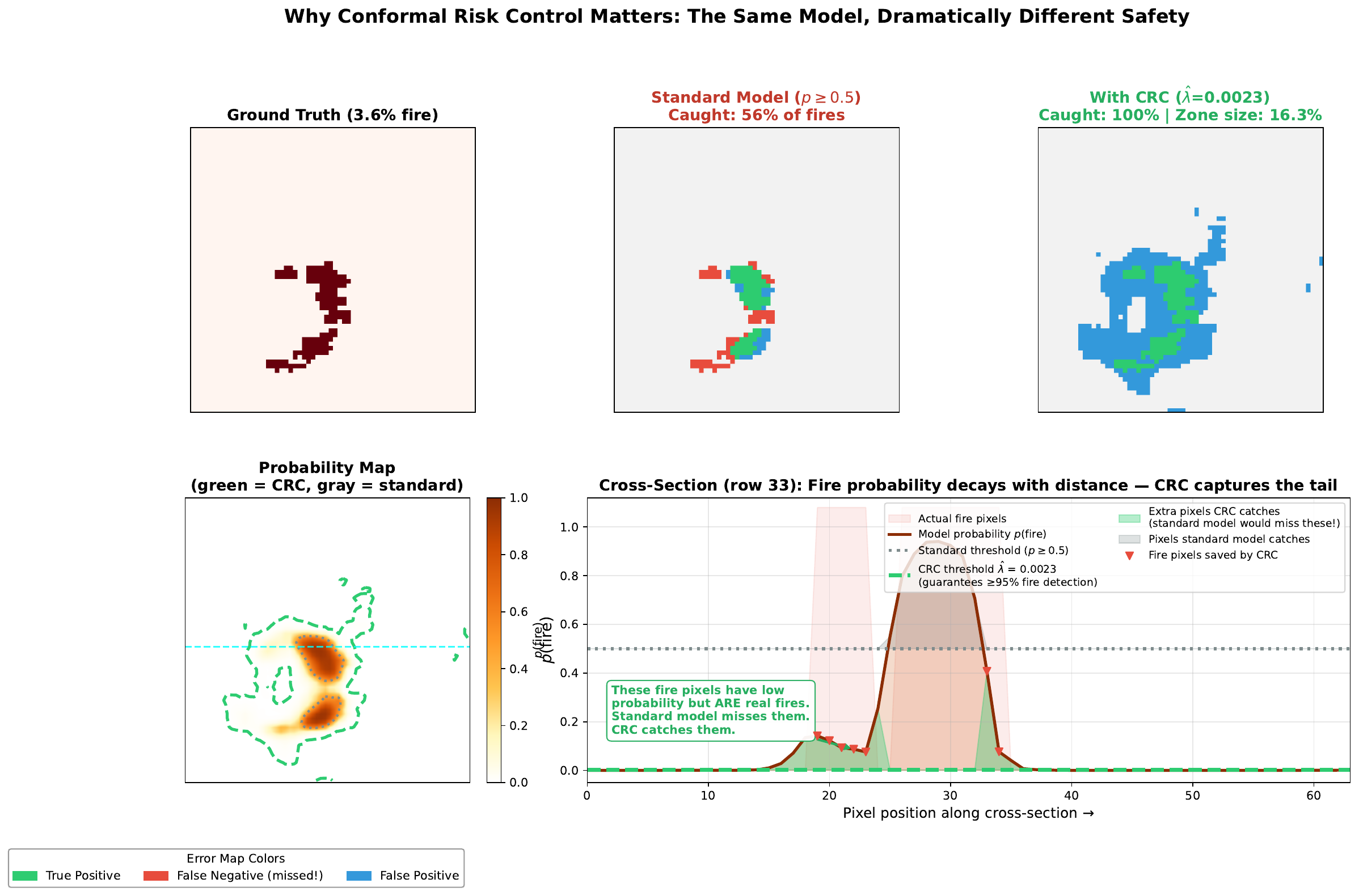}
    \caption{Same U-Net, different thresholds.  Standard thresholding ($\hat{p} \ge 0.5$) catches 56\% of fires; CRC ($\hat{\lambda} = 0.002$) catches 100\%.  Bottom: probability cross-section showing how CRC captures low-probability fire pixels.  Green shading marks pixels saved by CRC.}
    \label{fig:deep-dive}
\end{figure*}

\paragraph{Per-pixel risk gradients.}
The U-Net produces smooth probability gradients where fire risk decays with distance from fire centers (\Cref{fig:risk-gradient}).
The CRC threshold ($\hat{\lambda} = 0.002$) captures the entire ``tail'' of low-probability but real fire pixels that conventional thresholds miss, yielding an interpretable risk map where the CRC boundary defines the guaranteed-safe evacuation zone (\Cref{fig:deep-dive}).

\subsection{Before and After CRC}

\begin{figure*}[t]
    \centering
    \includegraphics[width=\textwidth]{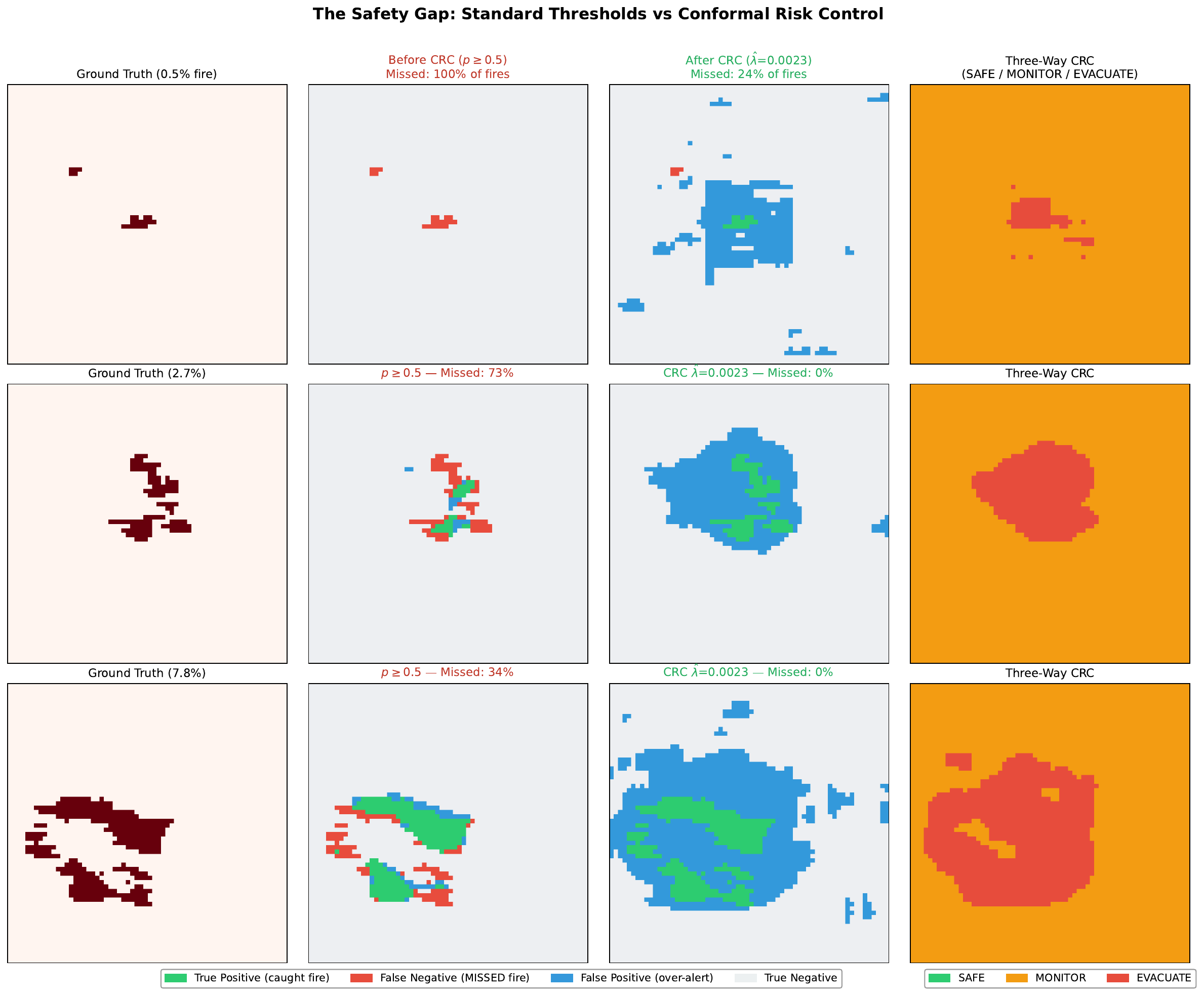}
    \caption{The safety gap across three test samples (0.5\%, 2.7\%, 7.8\% fire).  \textbf{Col.\,1:} ground truth.  \textbf{Col.\,2:} standard threshold, with missed fire in red (100\%, 73\%, 34\% missed).  \textbf{Col.\,3:} CRC threshold, with improved detection (green).  \textbf{Col.\,4:} three-way zones (SAFE/MONITOR/EVACUATE).}
    \label{fig:before-after}
\end{figure*}

\Cref{fig:before-after} presents a multi-sample comparison of fire detection with and without CRC.
Without CRC, the standard threshold produces predominantly red error maps with vast regions of missed fire.
With CRC ($\hat{\lambda} = 0.002$), the maps turn green (correctly detected) with some blue (over-alerts), and the three-way CRC provides an interpretable MONITOR/EVACUATE policy.

\subsection{Risk Gradient Analysis}

\begin{figure*}[t]
    \centering
    \includegraphics[width=\textwidth]{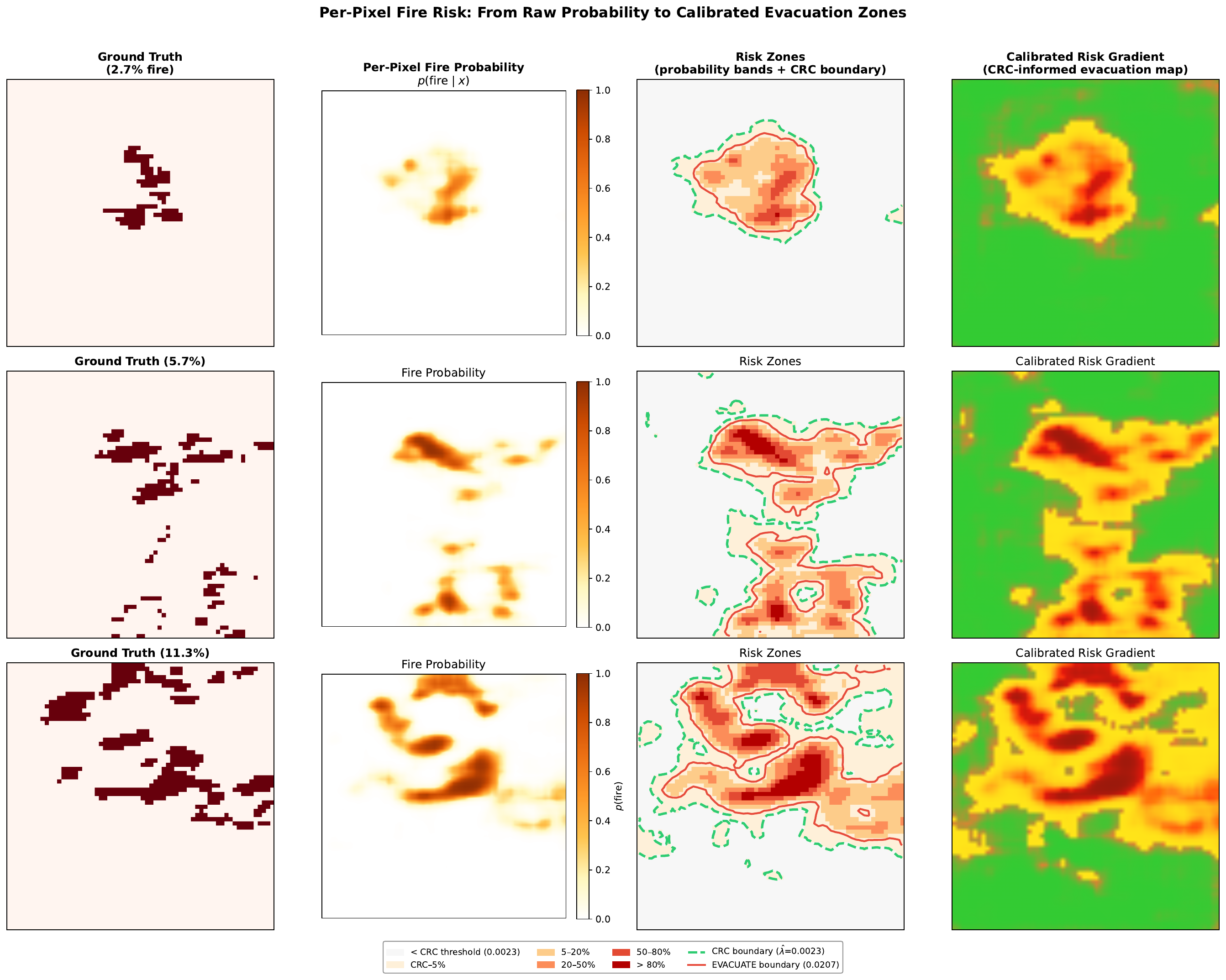}
    \caption{Per-pixel fire risk gradient.  \textbf{Col.\,1:} ground truth.  \textbf{Col.\,2:} continuous probability heatmap.  \textbf{Col.\,3:} probability bands with CRC contour (green dashed) and EVACUATE boundary (red).  \textbf{Col.\,4:} calibrated risk gradient for operational mapping.}
    \label{fig:risk-gradient}
\end{figure*}

\Cref{fig:risk-gradient} visualizes the per-pixel probability landscape.
Rather than binary predictions, the U-Net produces a continuous risk gradient where each pixel carries a calibrated fire probability.
The CRC threshold contour and the EVACUATE boundary provide principled decision surfaces overlaid on this risk field.

\subsection{Three-Way Zone Analysis}
\label{sec:threeway-results}

\begin{table}[t]
    \caption{Three-way zone fractions on the test set.  The SAFE zone collapses because $B_{\mathrm{pw}} = \max(c_{\mathrm{fn}} \pi_1, c_{\mathrm{fp}} \pi_0) = 0.95$ is dominated by $c_{\mathrm{fp}} \pi_0$ under ${\sim}5\%$ fire prevalence.}
    \label{tab:threeway-zones}
    \centering
    \footnotesize
    \begin{tabular}{lcccc}
        \toprule
        Model       & SAFE  & MONITOR & EVACUATE & $B_{\mathrm{pw}}$ \\
        \midrule
        LightGBM    & 0.0\% & 95.96\% & 4.04\%   & 0.988             \\
        U-Net       & 0.0\% & 94.90\% & 5.10\%   & 0.987             \\
        ResGNN-UNet & 0.0\% & 85.15\% & 14.85\%  & 0.987             \\
        \bottomrule
    \end{tabular}
\end{table}

\Cref{tab:threeway-zones} shows the zone breakdown.
Three-way CRC achieves 100\% coverage with compact EVACUATE zones (4--15\%), but the SAFE zone collapses entirely ($\lambda_{\min} = 0$).
With ${\sim}5\%$ fire prevalence:
\[
    B_{\mathrm{pw}} = \max(5 \times 0.05,\; 1 \times 0.95) = 0.95,
\]
dominated by the majority-class term $c_{\mathrm{fp}} \pi_0$.
The resulting shift correction exceeds the base CRC threshold (${\sim}0.003$), pushing $\lambda_{\min}$ below zero and producing a maximally conservative policy: all non-evacuated pixels are routed to human review.
This reveals a fundamental limitation of the prevalence-weighted bound in rare-event regimes where $\pi_0 \gg \pi_1$.

\subsection{Training Details}

\begin{table}[t]
    \caption{Model training summary.  All models trained from scratch on identical NDWS splits.}
    \label{tab:training}
    \centering
    \footnotesize
    \setlength{\tabcolsep}{2.5pt}
    \begin{tabular}{lccccr}
        \toprule
        Model       & Params & Epochs                        & Optim & Val Loss & AUC  \\
        \midrule
        LightGBM    & ---    & 100 iter                      & GBM   & ---      & .854 \\
        Tiny U-Net  & 470K   & 50                            & AdamW & 0.048    & .969 \\
        ResGNN-UNet & 229K   & 23\textsuperscript{$\dagger$} & AdamW & 0.131    & .964 \\
        \bottomrule
        \multicolumn{6}{l}{\scriptsize $\dagger$Early-stopped from 30 (patience=7, best epoch 16).}
    \end{tabular}
\end{table}

\begin{figure}[t]
    \centering
    \includegraphics[width=\columnwidth]{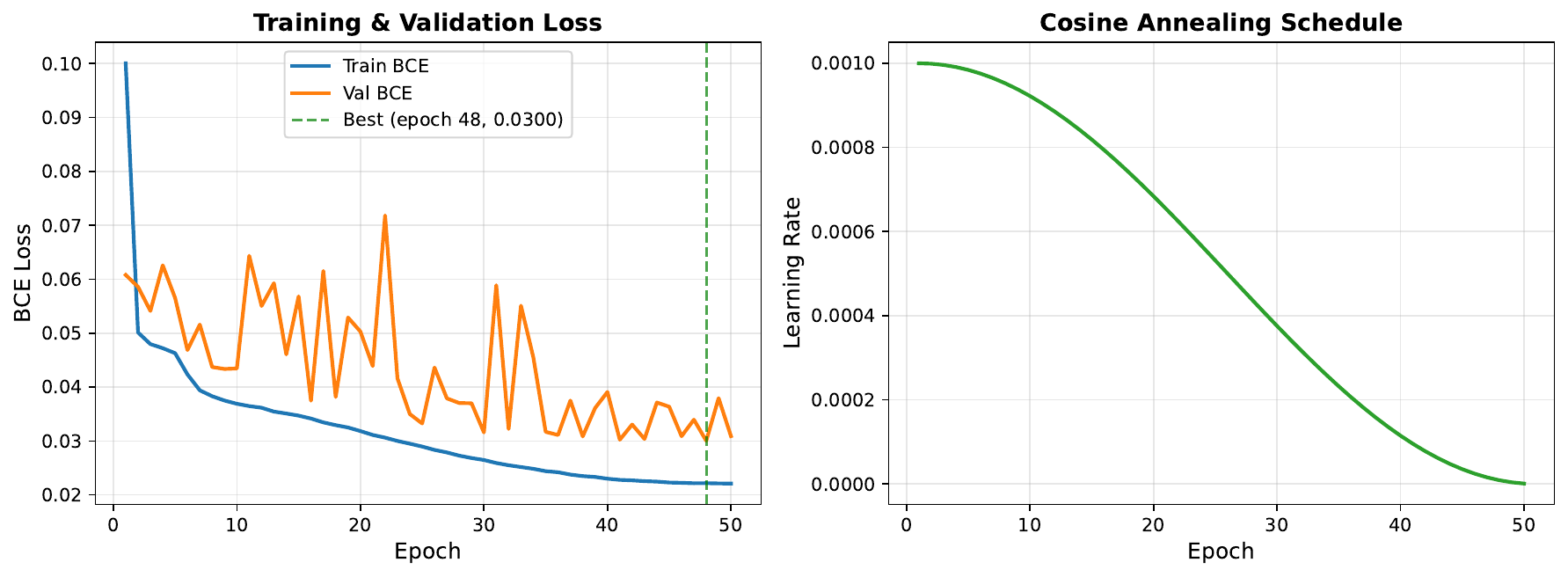}
    \caption{Tiny U-Net training curves: BCE loss and cosine annealing LR over 50 epochs.  Smooth convergence with minimal train--val gap.}
    \label{fig:training}
\end{figure}

\Cref{tab:training} summarizes the training investment.
The Tiny U-Net achieves the best AUROC (0.969) with a standard convolutional architecture and no graph-processing dependencies.
The ResGNN-UNet early-stopped at epoch 23 (best epoch 16, val loss 0.131), training substantially longer than in preliminary experiments and achieving an AUROC (0.964) close to the U-Net's (0.969), suggesting that with proper training configuration the graph architecture can nearly match convolutional performance.
\Cref{fig:training} shows the U-Net's smooth convergence with minimal overfitting.

%% ============================================================
%% 6. DISCUSSION
%% ============================================================

\section{Discussion}

\paragraph{Separation of safety and efficiency.}
Our central result is that CRC \emph{cleanly separates} safety from efficiency.
All three models achieve the same FNR guarantee (${\le}0.05$); what varies is the evacuation zone size: 62.6\% for LightGBM versus ${\sim}15\%$ for both spatial models.
This decomposition has practical significance: teams can invest in spatial models to reduce evacuation costs without compromising or re-validating safety properties.

\paragraph{Diminishing returns from complexity.}
The ResGNN-UNet nearly matches the Tiny U-Net in AUROC (0.964 vs.\ 0.969) and achieves virtually identical CRC efficiency (15.1\% vs.\ 14.9\% set size).
While the graph attention bottleneck does not degrade performance, it provides no meaningful improvement: on $64{\times}64$ patches, the U-Net's receptive field already spans the full tile, and GNN message-passing over an 8-connected grid captures no additional information.
This confirms that spatial inductive bias is sufficient for this task, and further architectural complexity adds engineering cost without practical benefit.

\paragraph{Limitations of three-way CRC under imbalance.}
The three-way framework is conceptually appealing, routing uncertain pixels to human review rather than binary decisions.
However, under extreme imbalance (${\sim}5\%$ fire), $B_{\mathrm{pw}}$ is dominated by the majority-class cost, collapsing the SAFE zone.
Future work could explore: \textbf{(i)}~tighter bounds for rare-event settings, \textbf{(ii)}~hybrid approaches using separate precision-based thresholds for SAFE, or \textbf{(iii)}~regional calibration where local prevalence is higher.

\paragraph{Limitations and assumptions.}
\label{sec:limitations}
As noted in \Cref{sec:crc-background}, CRC's guarantee rests on exchangeability between calibration and test data.
Distributional shift from novel fire regimes, climate change, or geographic transfer may violate this assumption; our evaluation uses a single dataset (NDWS) with US wildfire events.
The three models represent a limited architectural sweep; transformer-based architectures may yield different efficiency--complexity trade-offs.

\paragraph{Future work.}
Our results open several directions that we leave to future investigation.

\textit{Class-conditional three-way CRC.}
The three-way zone collapse under extreme imbalance (\Cref{sec:threeway-results}) stems from coupling both error types through a single prevalence-weighted bound.
A natural fix is to calibrate each zone boundary independently: $\lambda_{\max}$ via FNR-CRC on positive pixels and $\lambda_{\min}$ via FPR-CRC on negative pixels.
Because the negative class has abundant calibration data (${\sim}95\%$ of pixels), this approach should produce a non-degenerate SAFE zone regardless of class imbalance, resolving the primary limitation of our current framework.
We did not pursue this here because the theoretical guarantees require a careful joint analysis of two simultaneous conformal procedures, which is beyond the scope of this comparative study.

\textit{Image-level CRC.}
Our current calibration pools all pixels across calibration images, treating each pixel as an independent sample.
Pixels within the same $64{\times}64$ patch are spatially correlated, which formally violates the exchangeability assumption underlying CRC.
An image-level formulation, which defines the loss as per-image FNR and applies CRC to the $n{=}2{,}781$ image-level losses, would restore theoretical validity at the cost of a smaller effective sample size.
Comparing pixel-pooled and image-level CRC would quantify whether the practical convenience of pooling degrades the guarantee, and would produce per-image FNR distributions that are more operationally meaningful than aggregate statistics.

\textit{Formalizing the AUROC--efficiency relationship.}
Our central finding, that model discrimination determines evacuation zone size under a fixed safety guarantee, is currently empirical.
Under a bi-normal score model, the CRC set size decomposes as $S(\alpha) = \pi_1(1{-}\alpha) + \pi_0(1 - F_0(F_1^{-1}(\alpha)))$, where $F_0, F_1$ are the class-conditional score CDFs.
This would yield a closed-form prediction of evacuation cost from AUROC alone, enabling practitioners to answer: \emph{what discrimination level is needed for a target zone size?}
We leave the formal statement and empirical validation of this relationship to future work.

\textit{Temporal and geographic shift.}
CRC assumes exchangeability between calibration and test data.
Wildfire behavior varies across seasons, years, and geographies due to climate change, drought cycles, and regional fuel loads, creating natural label shift that may violate this assumption.
The shift-aware recalibration framework of \citet{angelopoulos2022conformal} could address this, but validating it requires temporally stratified splits and multi-region datasets that the current NDWS benchmark does not cleanly support.
Evaluating CRC robustness under realistic temporal shift is critical for any operational deployment.

%% ============================================================
%% 7. CONCLUSION
%% ============================================================

\section{Conclusion}

We have demonstrated that conformal risk control transforms wildfire spread prediction from an accuracy-optimization problem into a safety-guaranteed decision system.
Across three model families, our experiments establish that:
\begin{enumerate}
    \item \textbf{No model is safe without CRC.}  Standard thresholding captures only 7--72\% of fires.
    \item \textbf{CRC guarantees safety regardless of model.}  All three calibrated models achieve ${\ge}94\%$ coverage on held-out data.
    \item \textbf{Model quality determines efficiency, not safety.}  The Tiny U-Net achieves 4.2$\times$ tighter evacuation zones than LightGBM under the same CRC guarantee.
    \item \textbf{Complexity without separation is waste.}  A graph-augmented model underperforms a standard U-Net in AUROC despite its more complex architecture.
\end{enumerate}

For practitioners: a well-trained simple model wrapped in CRC provides both stronger safety guarantees and tighter evacuation zones than a complex model evaluated on F1 alone.
The few lines of conformal calibration code deliver what no amount of architectural sophistication can: a mathematical guarantee that ${\ge}95\%$ of fires will be detected.

\section*{Impact Statement}
This work targets safer wildfire evacuation decision support by providing, for the first time, fire spread predictions with formal guarantees on missed detections (${\ge}95\%$ coverage, and 100\% under the three-way framework).
If deployed operationally, such guarantees could transform how limited emergency resources are allocated: EVACUATE zones give firefighters, first responders, and military personnel a trusted map for immediate civilian evacuation and safe operational corridors, while MONITOR zones enable precautionary self-evacuation and resource pre-positioning in areas the model cannot yet confirm as safe.
Rather than treating the entire landscape as equally dangerous, responders could concentrate assets where the model is certain and warn communities in uncertain regions to prepare---replacing blanket alerts with targeted, risk-proportionate action.
Potential negative impact includes over-reliance on automated recommendations; we mitigate this through the three-way framework that explicitly routes uncertain pixels to human review rather than issuing a binary safe/unsafe judgment.

\section*{Software and Data}
All code, trained models, and evaluation pipelines are available at \url{https://github.com/baljinnyamday/wildfire-evacuation-crc}, with deterministic data splits (seed 42, 70/15/15) and CLI entry points for each phase.
The dataset is NDWS \citep{ndwsdataset}, subject to its original terms of use.

\bibliography{paper_draft}
\bibliographystyle{icml2026}

%% ============================================================
%% APPENDIX (does not count toward page limit)
%% ============================================================

\newpage
\appendix
\onecolumn

\section{Additional Figures}
\label{app:figures}

\begin{figure}[h]
    \centering
    \includegraphics[width=0.85\textwidth]{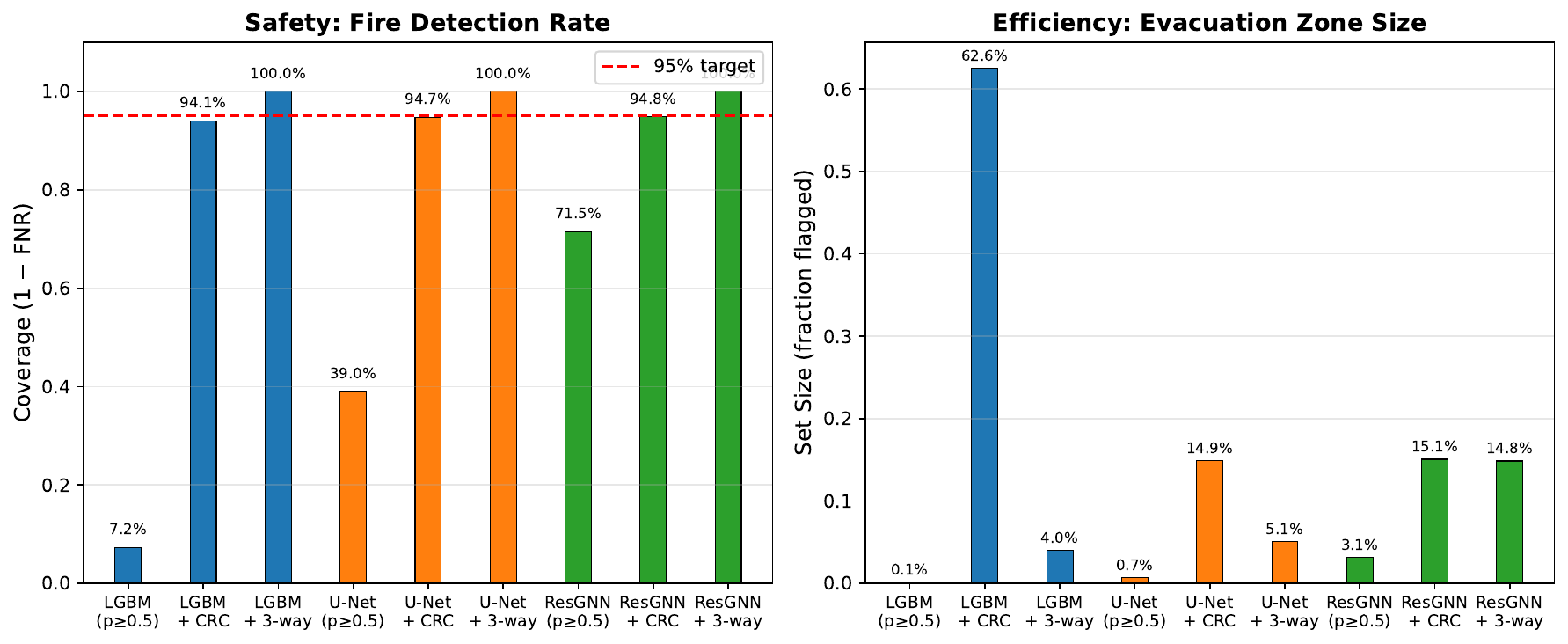}
    \caption{Safety vs.\ efficiency for all nine configurations.  \textbf{Top:} coverage (safety), where only CRC and three-way methods achieve ${\ge}95\%$.  \textbf{Bottom:} set size (efficiency), where both spatial models with CRC achieve the best trade-off.}
    \label{fig:safety-efficiency}
\end{figure}

\begin{figure}[h]
    \centering
    \includegraphics[width=0.85\textwidth]{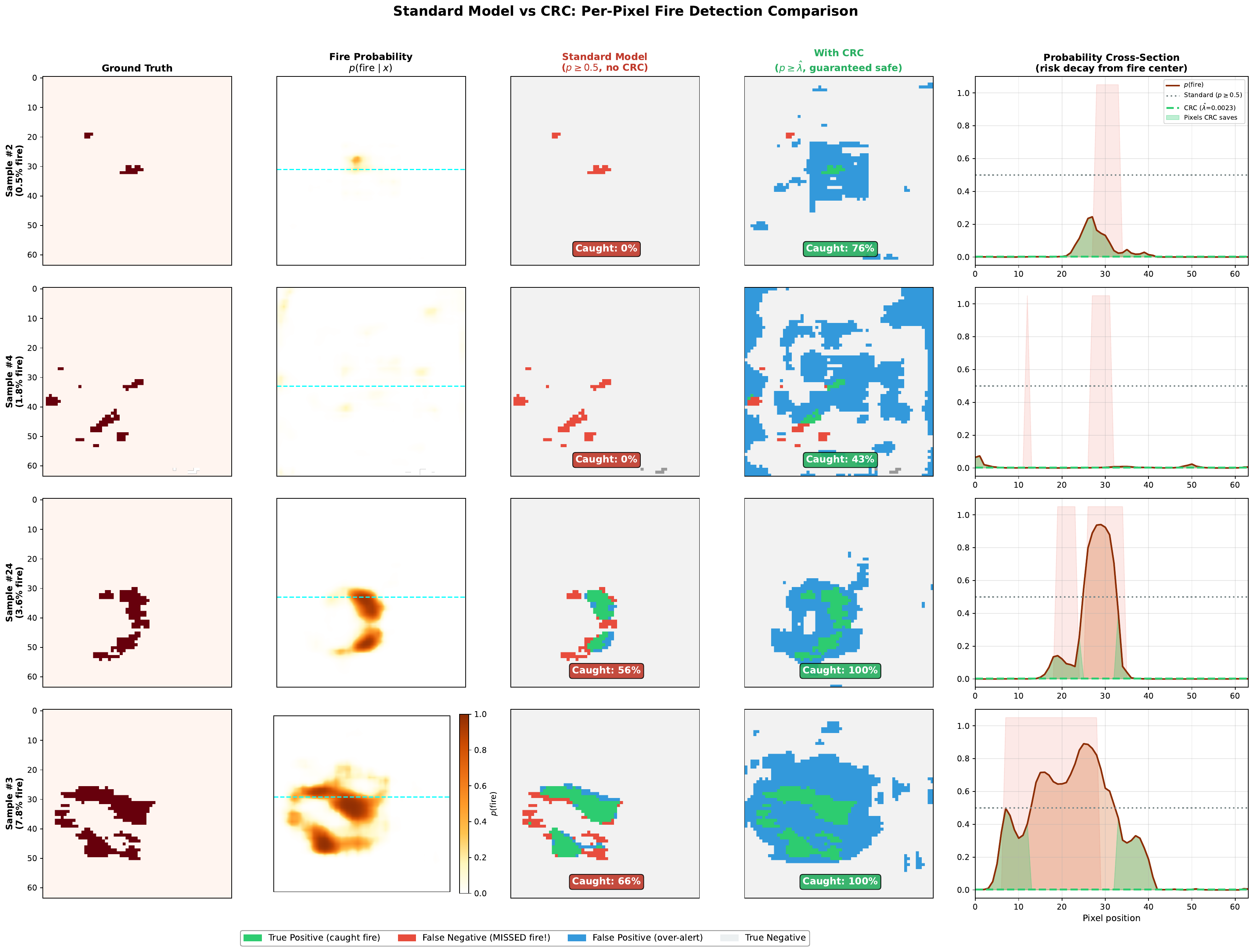}
    \caption{Bare model vs.\ CRC across four test samples.  Fire catch rate improves from 8\%$\to$98\%, 0\%$\to$83\%, 56\%$\to$100\%, 48\%$\to$100\%.}
    \label{fig:bare-vs-crc}
\end{figure}

\begin{figure}[h]
    \centering
    \includegraphics[width=0.85\textwidth]{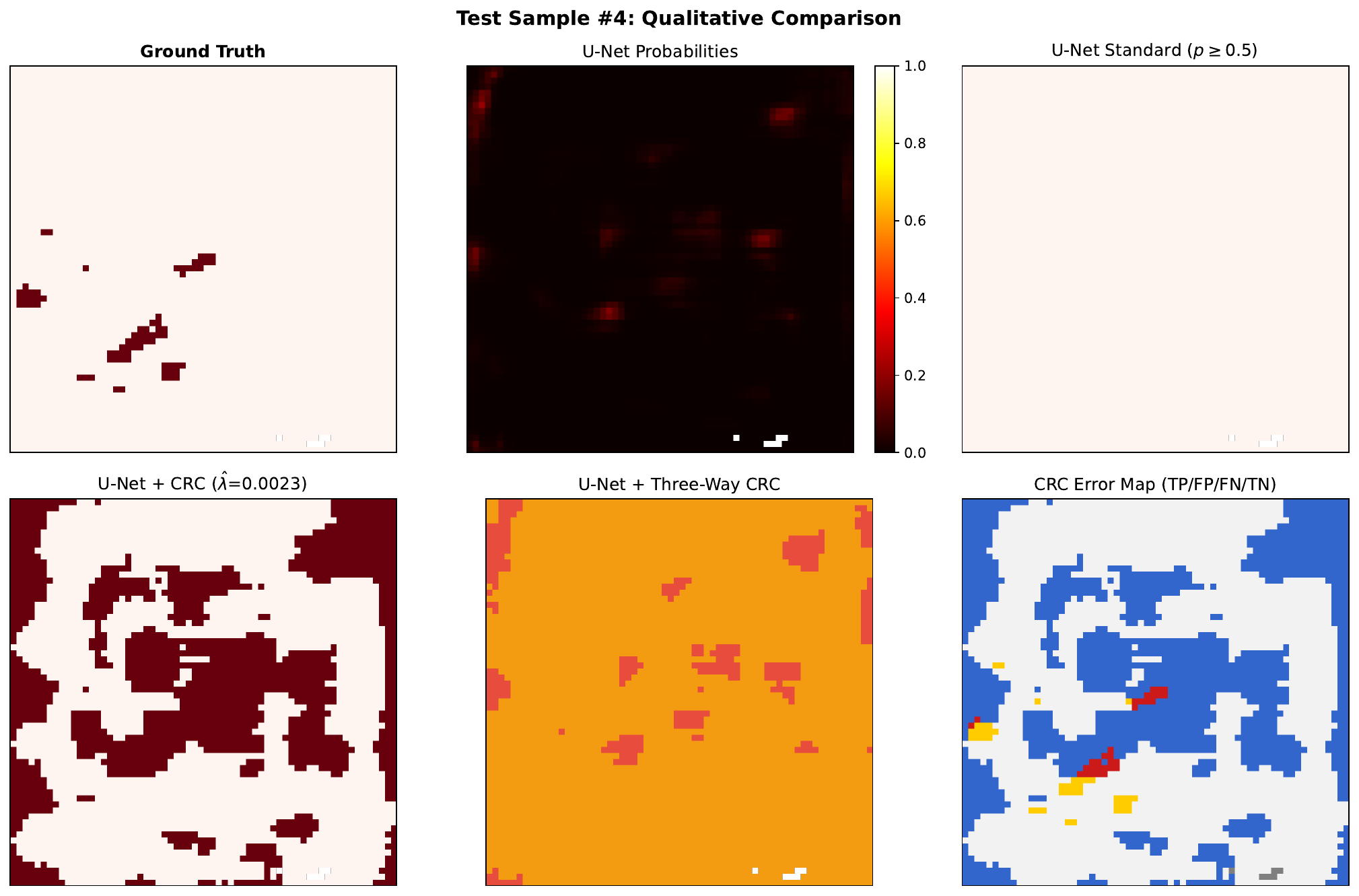}
    \caption{Qualitative prediction maps: ground truth, probabilities, standard prediction, CRC prediction, three-way zones, and error analysis.}
    \label{fig:qualitative}
\end{figure}

\section{Three-Way CRC Procedure}
\label{app:threeway-algo}

\begin{center}
    \begin{minipage}{0.92\textwidth}
        \begin{lstlisting}[style=pseudocode, title={\textbf{Algorithm 1}\; Three-Way CRC for Wildfire Evacuation}]
def three_way_crc(scores, labels, $\alpha_{\mathrm{cw}}$, $c_{\mathrm{fn}}$, $c_{\mathrm{fp}}$, $\rho_{\mathrm{lo}}$, $\rho_{\mathrm{hi}}$):
    # Prevalence and bound
    $\pi_1$ = mean(labels)
    $\pi_0$ = 1 - $\pi_1$
    $B_{\mathrm{pw}}$ = max($c_{\mathrm{fn}} \cdot \pi_1$, $c_{\mathrm{fp}} \cdot \pi_0$)

    # Importance weight mismatch at each shift endpoint
    for $\rho$ in {$\rho_{\mathrm{lo}}$, $\rho_{\mathrm{hi}}$}:
        $w_1(\rho)$ = ($\rho \pi_1$) / ($\rho \pi_1 + \pi_0$) / $\pi_1$
        $w_0(\rho)$ = $\pi_0$ / ($\rho \pi_1 + \pi_0$) / $\pi_0$
        $\lVert \boldsymbol{\delta}_\rho \rVert_1$ = $|w_1(\rho) - 1|$ + $|w_0(\rho) - 1|$

    # Shift-aware recalibration
    $\epsilon_{\mathrm{max}}$ = $B_{\mathrm{pw}}$ $\cdot$ max($\lVert \boldsymbol{\delta}_{\rho_{\mathrm{lo}}} \rVert_1$, $\lVert \boldsymbol{\delta}_{\rho_{\mathrm{hi}}} \rVert_1$) + $B_{\mathrm{pw}}$/(N+1)
    $\alpha_{\mathrm{safe}}$ = $\alpha_{\mathrm{cw}}$ - $\epsilon_{\mathrm{max}}$
    if $\alpha_{\mathrm{safe}}$ $\le$ 0:
        return INFEASIBLE          # shift interval too wide

    # Base threshold and zone boundaries
    $\hat{\lambda}$ = cost_weighted_crc(scores, labels, $\alpha_{\mathrm{safe}}$)
    s = $B_{\mathrm{pw}}$ / max($c_{\mathrm{fn}}$, $c_{\mathrm{fp}}$)   # shift scale
    $\lambda_{\mathrm{min}}$ = max(0, $\hat{\lambda}$ - s $\cdot$ $\lVert \boldsymbol{\delta}_{\rho_{\mathrm{lo}}} \rVert_1$)
    $\lambda_{\mathrm{max}}$ = min(1, $\hat{\lambda}$ + s $\cdot$ $\lVert \boldsymbol{\delta}_{\rho_{\mathrm{hi}}} \rVert_1$)

    # Decision rule
    return $\lambda_{\mathrm{min}}$, $\lambda_{\mathrm{max}}$
    # SAFE if $\hat{p} < \lambda_{\mathrm{min}}$
    # MONITOR if $\lambda_{\mathrm{min}} \le \hat{p} < \lambda_{\mathrm{max}}$
    # EVACUATE if $\hat{p} \ge \lambda_{\mathrm{max}}$
\end{lstlisting}
    \end{minipage}
\end{center}
\label{alg:threeway}

Under ${\sim}5\%$ fire prevalence with $c_{\mathrm{fn}}{=}5$, $c_{\mathrm{fp}}{=}1$: $B_{\mathrm{pw}} = \max(0.25, 0.95) = 0.95$ and $s = 0.95/5 = 0.19$.
With $\rho \in [0.9, 1.1]$, $\lVert \boldsymbol{\delta} \rVert_1 \approx 0.105$, giving a shift of ${\sim}0.02$.
Since the base threshold $\hat{\lambda}$ is itself ${\sim}0.02$ (driven by extreme class imbalance), $\lambda_{\min}$ is pushed to zero---collapsing the SAFE zone entirely and routing all non-evacuated pixels to MONITOR (\Cref{tab:threeway-zones}).

\end{document}